\documentclass[11pt,a4paper]{article}
\usepackage{times,latexsym}
\usepackage{url}
\usepackage[T1]{fontenc}

\usepackage[acceptedWithA]{tacl2021v1}

\usepackage{xspace,mfirstuc,tabulary}

\newif\iftaclinstructions
\taclinstructionsfalse 
\iftaclinstructions
\renewcommand{\confidential}{}
\renewcommand{\anonsubtext}{(No author info supplied here, for consistency with
TACL-submission anonymization requirements)}
\newcommand{\instr}
\fi

\iftaclpubformat 

\else

\fi

\usepackage{amsfonts}
\usepackage{amsmath}
\usepackage{array}
\usepackage{bbm}
\usepackage{booktabs}
\usepackage{graphicx}
\usepackage{multirow}
\usepackage{ragged2e}

\definecolor{bblue}{rgb}{0.0, 0.6, 0.8}
\definecolor{tab:blue}{rgb}{0.25, 0.617, 0.871}
\definecolor{tab:orange}{rgb}{1.0, 0.624, 0.291}
\definecolor{tab:green}{rgb}{0.294, 0.806, 0.294}

\usepackage{enumitem}
\setlist{nosep,leftmargin=*}

\usepackage[table]{xcolor}

\title{Rewriting History: A Recipe for Interventional Analyses \\ to Study Data Effects on Model Behavior}

\author{
Rahul Nadkarni$^1$ \quad
Yanai Elazar$^{2}$\Thanks{Equal contribution, work done while at the University of Washington.} \quad
Hila Gonen$^{3}$\footnotemark[1] \quad
Noah A.~Smith$^{1,4}$  \\
$^1$Paul G.~Allen School of Computer Science \& Engineering, University of Washington \\
$^2$Department of Computer Science, Bar-Ilan University \\
$^3$Department of Computer Science, University of British Columbia \\
$^4$Allen Institute for Artificial Intelligence
  \\  \texttt{\{rahuln,nasmith\}@cs.washington.edu}
  \\  \texttt{\{yanaiela,hilagnn\}@gmail.com}
}

\date{}

\begin{document}
\maketitle

\begin{abstract}
We present an experimental recipe for studying the relationship between training data and language model (LM) behavior. We outline steps for intervening on data batches -- i.e., ``rewriting history'' -- and then retraining model checkpoints over that data to test hypotheses relating data to behavior. Our intervention recipe's stages are (1) selecting evaluation items from a benchmark that measures model behavior, (2) matching relevant documents to those items, and (3) modifying those documents before retraining and measuring the effects. We demonstrate the utility of our recipe through case studies on factual knowledge acquisition and gender bias in LMs, using both cooccurrence statistics and information retrieval methods to identify documents that might contribute to model behavior. Our results supplement past observational analyses that link cooccurrence to model behavior, while demonstrating that extant methods for identifying relevant training documents do not fully explain an LM's abilities and biases. Researchers can follow the recipe to test further hypotheses about how training data affects model behavior. Our code is made publicly available to promote future work.\footnote{\url{https://github.com/rahuln/pretrain-intervention}.}
\end{abstract}

\section{Introduction}

When training LMs, the choice of pretraining data is generally believed to have a profound effect on what the learned models can do and how well they can do it. Numerous prior works have curated large pretraining datasets and studied the effect of pretraining data composition on downstream tasks \cite{li2024datacomp, longpre2024}, with some explicitly predicting downstream task performance \cite{grattafiori2024llama3herdmodels, bhagia2024establishingtaskscalinglaws, magnusson2025datadecide}. The primary motivation here has generally been performance improvement, not necessarily understanding model behavior, while mainly operating on the scale of entire pretraining datasets.

An equally important body of work relates pretraining data to model behavior to understand \emph{how} data affects performance without necessarily optimizing for it, retraining over modified data to do so. This includes various data-behavior relationships, such as how relative frequencies of verb tenses change subject-verb agreement performance \cite{wei2021} or how pronouns affect gender bias \cite{biderman2023}. Some recent work focuses on the specific behavior of factual knowledge acquisition, retraining models to understand how data gives rise to knowledge learning. This work operates across a range of scales, from pretraining from scratch on fictional biographies \citep{allen-zhu2024, allen-zhu2025, zucchet2025}, to continued pretraining of a base model on carefully-constructed documents expressing factual relations \citep{zhang2024}, and even intervening on intermediate stages of pretraining by injecting synthetic data to study the effect of repetition and paraphrasing \cite{howdollms2024} or entropy across activations \cite{kim2025} on fact learning and forgetting.

\begin{figure*}[t!]
    \centering
    \includegraphics[width=0.9\textwidth]{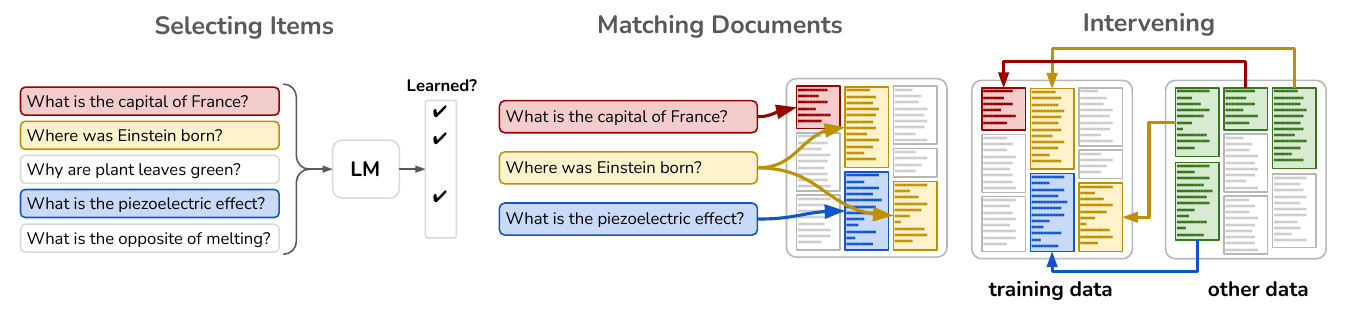}
    \caption{Diagram illustrating our proposed interventional analysis recipe, with stages including selecting items, matching documents to items, and intervening on matched documents.}
    \vspace{-.4cm}
    \label{fig:figure1}
\end{figure*}

While the aforementioned works all modify pretraining data to study model behavior, each takes a specific approach that does not easily transfer to other hypotheses. Seeking to draw connections across these works and formalize their methodologies, we propose a general recipe for \emph{interventional analyses} to study the relationship between pretraining data and model behavior at a fine-grained level through \emph{retraining} on modified data. Our proposed recipe (illustrated in Fig.~\ref{fig:figure1}) consists of three stages: (1) selecting which items in a benchmark to intervene on; (2) matching pretraining documents to those items; and (3) modifying those documents before repeating that portion of pretraining and measuring the effects. Our recipe organizes and comprehensively articulates the different considerations one must take into account when performing such an intervention, contextualized within past work. Our goal is to provide a recipe for practitioners to use when testing hypotheses about how specific properties of pretraining documents give rise to model behaviors, allowing for flexibility in the scale of the intervention (how much data is changed) and what portion of the training process is studied.

We demonstrate the utility of our recipe through case studies on \emph{factual knowledge acquisition} and \emph{gender bias}. LMs have been established as sources of factual information that can be queried with natural language \cite{petroni-etal-2019-language, roberts-etal-2020-much}, but how pretraining data yields knowledge capabilities is still under study. Similarly, LMs are known to exhibit various forms of bias \cite{navigli2023llmbias} including gender bias \cite{bordia-bowman-2019-identifying, nadeem-etal-2021-stereoset, van-der-wal-etal-2022-birth} that can be linked to their training data, yet direct interventions that tie bias to data are limited \cite{biderman2023}. We perform three case studies to explore these phenomena: the first relates term cooccurrence to fact triple completion (\S\ref{sec:pararel-results}), the second links pronoun-profession frequency and gender bias in coreference resolution (\S\ref{sec:gender-bias-results}), and the third uses information retrieval methods to study more complex knowledge queries (\S\ref{sec:mcqa-results}). These studies aim to improve our understanding of phenomena like knowledge acquisition and gender bias through intervening on and retraining over pretraining data, complementing previous observational studies while also showcasing our recipe's utility for different data sources and hypotheses.

For our first analysis, we operationalize knowledge acquisition through $\langle$subject, relation, object$\rangle$ fact triplets such as $\langle$``France'', ``capital city'', ``Paris''$\rangle$, considering a fact as having been learned when a model can identify the object given the subject and relation, i.e., correctly select ``Paris'' from a list of candidates when prompted with ``The capital city of France is''. We use our recipe to study the hypothesis that fact learning is related to subject-object \emph{cooccurrence} in pretraining documents. Cooccurrence has long been used as a source of distant supervision for aligning facts to text \cite{mintz-etal-2009-distant, augenstein2016, elsahar-etal-2018-rex} and has a well-established relationship to LM behavior through \emph{observational} analyses on the scale of entire pretraining datasets and fully-pretrained models \cite{termfreq2022, longtail2023, measuringcausal2022, kang2023, zhang2024, mallen2023, merullo2025linear, wanggeneralization, Verma2024, seshadri2024bias}.

The second study applies the recipe to better understand how pretraining data leads to gender bias in a coreference resolution task, where models exhibit better performance when certain gendered pronouns are paired with their stereotyped professions \cite{zhao-etal-2018-gender, rudinger-etal-2018-gender}. We again explore the effect of term cooccurrence, by swapping the gender of pronouns in documents that mention target professions to mitigate bias \cite{biderman2023}. By analyzing another kind of model behavior, this study illustrates the generality of our recipe.

Finally, we apply our recipe to multiple-choice question answering (MCQA) benchmarks used to evaluate LM factual knowledge (MMLU, \citealp{hendrycks2021measuring}; OpenBookQA, \citealp{mihaylov-etal-2018-suit}; SciQ, \citealp{welbl-etal-2017-crowdsourcing}), exploring another way of expressing and querying for factual knowledge that requires moving beyond cooccurrence and further serves to demonstrate the flexibility of our recipe. Overall, our results advance the study of how pretraining documents give rise to model behavior through direct intervention, complementing post-hoc observational analysis.
\section{Overview of Recipe}
\label{sec:overview}

\subsection{Components}
\label{sec:components}

We begin with the components of our interventional analysis recipe for studying the relationship between pretraining\footnote{We focus only on pretraining in this work, but a similar recipe could be applied to other stages of training as well.} data and model behavior:

\begin{itemize}
    \item A pretrained LM $\mathcal{M}$. We assume access to a set of \textit{checkpoints} $\boldsymbol{\mathcal{M}} = \{ \mathcal{M}_1, \dots, \mathcal{M}_T \}$, where each checkpoint $\mathcal{M}_s$ consists of model weights at intermediate step $s$ during pretraining and $\mathcal{M}_T = \mathcal{M}$ is the final pretrained model. Since we wish to retrain from intermediate checkpoints, we also assume access to the optimizer state at each checkpoint.
    
    \item The pretraining dataset $\mathcal{D} = \{d_1, \dots, d_N\}$ used to train $\mathcal{M}$, where each $d_i$ is a sequence of tokens that form a document.\footnote{Note that each $d_i$ does not have to be a complete document from the original text data, as documents can be split and concatenated together to fit into the maximum sequence length window when training the LM.} We assume the dataset can be partitioned into $T$ \emph{data batches} $\mathcal{D}_1, \dots, \mathcal{D}_T$, such that $\mathcal{D}_s$ is the ordered sequence of tokens seen between checkpoints $\mathcal{M}_{s-1}$ and $\mathcal{M}_s$ during pretraining.
    
    \item A task or skill of interest represented by an evaluation dataset $\mathcal{E} = \{(x_1, y_1), \dots, (x_n, y_n)\}$ of \emph{test items}, where each item $(x_i, y_i)$ is an input and its corresponding correct output.

    \item An \emph{evaluation metric} $\mathcal{F}(\mathcal{M}_s, (x_i, y_i))$ that applies a checkpoint $\mathcal{M}_s$ to an item $(x_i, y_i)$ to quantify model behavior on that item (e.g., correctness at predicting $y_i$ given $x_i$, log-probability of $y_i$ given $x_i$, etc.).
\end{itemize}

Given $\mathcal{D}$ and a full specification of the LM training procedure, we can replicate pretraining from a model checkpoint $\mathcal{M}_s$ to $\mathcal{M}_{s+k}$ for any two checkpoints indexed $s$ and $s+k$, by retraining over the ordered data batches $\mathcal{D}_{s+1}, \dots, \mathcal{D}_{s+k}$. We can also make targeted interventions on those data batches and re-run pretraining to generate a counterfactual model checkpoint $\mathcal{M}_{s+k}^{\text{int}}$, which we propose doing in our interventional analysis recipe.

\subsection{Designing an Experiment}

Suppose we are interested in how pretraining data affects the behavior of our model $\mathcal{M}$ at some timestep $t$, as measured by applying our metric $\mathcal{F}$ to $\mathcal{M}_t$ on some target subset of test items $\mathcal{E}^{\text{target}} \subseteq \mathcal{E}$. We have some hypothesis about how the relationship between documents in $\mathcal{D}_t$\footnote{Here we consider only data seen since the previous model checkpoint, but in practice the partitioning of $\mathcal{D}$ into data batches can be defined in such a way that $\mathcal{D}_t$ spans multiple checkpoints.} and items in $\mathcal{E}^{\text{target}}$ cause this behavior, and we wish to intervene on $\mathcal{D}_t$ to test our hypothesis. Our interventional analysis recipe can be broken down into three stages:

\begin{itemize}
    \item Selecting a subset $\mathcal{E}^{\text{target}} \subseteq \mathcal{E}$ of target evaluation items on which our LM exhibits the behavior at timestep $t$, and corresponding data batch $\mathcal{D}_t$ to intervene on (\S\ref{sec:select-items});
    \item Identifying a set of relevant documents in $\mathcal{D}_t$ by matching to target items from $\mathcal{E}^{\text{target}}$ (\S\ref{sec:match-docs1});
    \item Intervening by modifying relevant documents in $\mathcal{D}_t$ to construct intervention data batch $\mathcal{D}_t^{\text{int}}$ and retraining over it starting from $\mathcal{M}_{t-1}$ to yield intervention checkpoint $\mathcal{M}_t^{\text{int}}$ (\S\ref{sec:modify-docs}).
\end{itemize}

Finally, we compare $\mathcal{M}_t$ and $\mathcal{M}_t^{\text{int}}$ by applying $\mathcal{F}$ to measure their behavior on $\mathcal{E}^{\text{target}}$, determining if the intervention confirms our hypothesis.

Since we are performing an interventional analysis, our recipe instantiates  \emph{causal inference}, relating properties of pretraining data to model behavior. Foundational texts on causality \cite{pearl2009causality} and its application in NLP \cite{feder-etal-2022-causal}  define a causal relationship between some property of the data (e.g., term cooccurrence) and some model behavior (e.g., task accuracy) through the \emph{causal effect} of that property, i.e., a measurable change in behavior after modifying the data through a controlled intervention, where the magnitude of the change reflects the strength of the causal relationship. Note that this definition (1) does not require that the modified property of the data is the \emph{sole} cause of the behavior, and (2) does not depend on the behavior being altered in some \emph{complete} way (e.g., accuracy dropping to zero). We do not expect or claim that our interventions target the sole causal factor in each case study, and in fact our results show that they do not. We present our recipe as a methodological tool, enabling future work that explores other potential causes of the kinds of model behavior that we study.

There are technical decisions to be made at each stage, and these will depend on the specific model behaviors one wishes to study, the hypotheses, and the computational budget. We go into detail for each of these stages below, with examples of how they can be implemented. Note that we do not provide an exhaustive overview of how to implement each stage of the analysis or anticipate all possible interventions.  We believe creative  choices will be essential in future studies, and offer a starting point of various options and their tradeoffs.  Exploration of other ways of implementing each stage will lead to new instantiations of this recipe that study a wider range of questions.

\subsection{Stage 1: Selecting Items and a Data Batch}
\label{sec:select-items}

Given an hypothesis about how data influences some model behavior of interest, we need to first select items from our evaluation dataset $\mathcal{E}$ on which our model $\mathcal{M}$ exhibits that behavior. For example, if we are interested in factual knowledge acquisition, we might select questions that our model answers correctly, whereas if we are interested in forgetting behavior we might select items that the model answers incorrectly at a given checkpoint after being correct for some number of previous checkpoints.

Formally, we wish to select a \emph{target subset} $\mathcal{E}^{\text{target}}$ of items from our evaluation dataset $\mathcal{E}$ (i.e., $\mathcal{E}^{\text{target}} \subseteq \mathcal{E}$). Consider a function of each item, all model checkpoints, and a particular timestep of the form
    $ f_\text{sel}( (x_i, y_i), \boldsymbol{\mathcal{M}}, s ) \in \{0, 1\} $,
which is $1$ when our model exhibits the behavior of interest on item $(x_i, y_i)$ at step $s$, and $0$ otherwise (note that this is a function of all model checkpoints $\boldsymbol{\mathcal{M}}$ since we may be interested in behavior that involves multiple checkpoints, such as the forgetting example above). Then we define the target subset $\mathcal{E}_s^{\text{target}} \subseteq \mathcal{E}$ of items at a given step $s$ as
    $$ \mathcal{E}_s^{\text{target}} = \{ (x_i, y_i) \in \mathcal{E} \mid f_\text{sel}((x_i, y_i), \boldsymbol{\mathcal{M}}, s ) = 1\} $$
Since we have different target subsets $\mathcal{E}_s^{\text{target}}$ for each step $s$, we need to choose a step to intervene on. This can be motivated by different considerations -- the stage of pretraining one is interested in, properties of items in $\mathcal{E}_s^{\text{target}}$, etc. One goal may be for  $\mathcal{E}^{\text{target}}$ to contain as many items as possible, as a larger sample size improves our ability to measure the true effect of our intervention \cite{card-etal-2020-little}. This may motivate choosing a target subset and corresponding timestep $t$ as the subset $\mathcal{E}_s^{\text{target}}$ of the largest size
    $$ t = \underset{s \; \in \; [1, \dots, T]}{\text{argmax}} \; \left\lvert \mathcal{E}_s^{\text{target}} \right\rvert \quad\quad \mathcal{E}^{\text{target}} = \mathcal{E}_t^{\text{target}}$$
There are also many ways to design $f_{\text{sel}}(\cdot)$ to select items, and this choice should be guided by the behavior under study. For example, \citet{howdollms2024} examine changes in log-probability of target spans in text probes to study factual knowledge acquisition, and \citet{porada2022} do the same for commonsense knowledge triples. Selection could also be based on intermediate layer activations: \citet{kim2025} track the entropy of activations in feedforward layers and relate it to fact learning and forgetting.

Existing work typically uses synthetic data or all items in a dataset rather than just ones on which the model exhibits a certain behavior, but we propose selecting a target subset since not all items in a benchmark may be relevant to the hypothesis (e.g., for fact learning, not all items may be learned at a given checkpoint). We believe properly selecting a target set of items is an important consideration when performing these kinds of analyses, and so we include it in our recipe for completeness.

\subsection{Stage 2: Identifying Relevant Documents}
\label{sec:match-docs1}

Given a target subset of items $\mathcal{E}^{\text{target}}$ on which our model exhibits the behavior of interest and intervention timestep $t$, we next identify a subset of pretraining documents within the data batch $\mathcal{D}_t$ that we hypothesize to affect the behavior on the target items. The matching process should also be motivated by the original hypothesis, i.e., what relationship between documents and items is believed to be driving model behavior.

In practice, it may be hard to identify documents that we believe contribute to model behavior on our target set, so we approximate with some heuristic for matching documents to items. For example, if our hypothesis is that pretraining documents which contain multiple terms that also appear in an item lead to the model making a correct prediction on that item, we should match the item to documents with some lexical overlap (e.g., term cooccurrence). Suppose we define a matching function
    $ g_\text{match}(d, (x_i, y_i)) \in \{ 0, 1 \}$, 
which has value $1$ if document $d$ matches item $(x_i, y_i)$, and $0$ otherwise.\footnote{This formulation can be general -- if we instead have a method that assigns a real-valued similarity score to each document-item pair quantifying their relatedness, we can threshold this score to produce an indicator function $g_\text{match}(\cdot)$ as described.} We can then define a function that matches a document to our entire target subset of items {$\mathcal{E}^{\text{target}}$} as
    $$ f_\text{match}( d, \mathcal{E}^{\text{target}} ) = \bigvee_{(x_i, y_i) \; \in \; \mathcal{E}^{\text{target}}} g_\text{match}( d, (x_i, y_i) ), $$
i.e., we consider a document $d$ to be matching if it matches \emph{any} item in our target group $\mathcal{E}^{\text{target}}$. We will use this function to identify documents that we wish to modify to perform our intervention.

There are several considerations when choosing a method of matching documents to items, including tradeoffs in how effective different methods are at identifying target documents. While term cooccurrence can be used as an effective proxy for expression of a relation between entities \cite{elsahar-etal-2018-rex}, spurious matches are possible (e.g., documents where ``Paris'' and ``France'' cooccur without expressing the relation ``The capital city of France is Paris''). Similarly, \citet{yauney2023} find that item-document similarity measured using n-gram distributions or pretrained model embeddings does not correlate with performance across a range of tasks, though without performing any interventions. Our recipe provides a way of testing the hypotheses that motivate using each of these methods to identify relevant documents.

Selecting a data batch to intervene on and even defining how much data constitutes a ``data batch'' in one's own analysis also requires careful consideration. While frequent checkpoints can lead to data batches with several million documents and enable a wide range of possible matching methods, applying our recipe to large batches or entire pretraining datasets with billions of documents would render all but the simplest of methods (e.g., searching for terms) computationally intractable. Thus, potential computational constraints may determine the scale of the intervention.

\subsection{Stage 3: Intervening}
\label{sec:modify-docs}

Our final stage performs the intervention by modifying the documents in $\mathcal{D}_t$ that match items in $\mathcal{E}^{\text{target}}$ and retraining over the modified $\mathcal{D}_t$ to observe the effects. Our intervention should also be motivated by the hypothesized relationship between documents in $\mathcal{D}_t$ and items in $\mathcal{E}^{\text{target}}$ such that it alters (e.g., removes) this relationship. As an example, if we believe that a model correctly answers ``What is the capital of France?'' with ``Paris'' because it was trained on  documents where ``Paris'' and ``France'' cooccur, it is sufficient to intervene on documents in any way that removes this cooccurrence. This can be as small an intervention as editing a few tokens, or as large as replacing the entire document with one in which ``Paris'' and ``France'' do not cooccur.

We define an intervention function $g_\text{int}(\cdot)$ that takes in a matched document, the set of target items, and the remaining pretraining data and returns a modified document 
    $ d' = g_\text{int}(d, \mathcal{E}^{\text{target}}, \mathcal{D})$.
We include $\mathcal{E}^{\text{target}}$ and $\mathcal{D}$ as arguments to the intervention function since we may want to make modifications that make use of properties of items (e.g., salient terms that occur in $x_i$ or $y_i$) or replace $d$ with other documents in $\mathcal{D}$. Since we only want to intervene on documents which match items in our target set $\mathcal{E}^{\text{target}}$, we define a function
\begin{equation*}
\resizebox{.9\hsize}{!}{$ f_\text{int}(d) =
\begin{cases}
    g_\text{int}(d, \mathcal{E}^{\text{target}}, \mathcal{D}) & \mbox{if } f_\text{match}(d, \mathcal{E}^{\text{target}} ) = 1 \\
    d & \mbox{if }f_\text{match}(d, \mathcal{E}^{\text{target}} ) = 0
\end{cases} $}
\end{equation*}

Then we construct our intervention data batch $\mathcal{D}_t^{\text{int}}$ from our original data batch $\mathcal{D}_t$ as
    $ \mathcal{D}_t^{\text{int}} = \{ f_\text{int}(d) \mid d \in \mathcal{D}_t \} $.
Finally, we re-run pretraining over the intervention data. We initialize with model checkpoint $\mathcal{M}_{t - 1}$ (and its optimizer state) and train over $\mathcal{D}_t^{\text{int}}$ instead of $\mathcal{D}_t$, resulting in an intervention checkpoint $\mathcal{M}_t^{\text{int}}$. We then apply our evaluation metric $\mathcal{F}$ to items in $\mathcal{E}^{\text{target}}$ using both $\mathcal{M}_t$ and $\mathcal{M}_t^{\text{int}}$ separately to compare their behavior and determine if the effect of our intervention is consistent with our hypothesis. The implementation depends on the evaluation metric. For instance, we can compare $\mathcal{F}(\mathcal{M}_t, (x_i, y_i))$ to $\mathcal{F}(\mathcal{M}_t^{\text{int}}, (x_i, y_i))$ for each $(x_i, y_i) \in \mathcal{E}^{\text{target}}$ (e.g., change in log-probability of $y_i$ given $x_i$), or we can aggregate $\mathcal{F}(\mathcal{M}_t, (x_i, y_i))$ over all $(x_i, y_i) \in \mathcal{E}^{\text{target}}$ (and similarly for $\mathcal{M}_t^{\text{int}}$) and compare the aggregated values (e.g., accuracy on $\mathcal{E}^{\text{target}}$).

The intervention function $g_{\text{int}}(\cdot)$ can vary in the degree to which it alters the pretraining data, depending on the hypothesis being studied. \citet{biderman2023} perform a token-level intervention that swaps masculine and feminine pronouns late in pretraining to observe the effects on gender bias, while \citet{wei2021} add and remove small subsets of documents to see how verb frequencies affect subject-verb agreement performance. At the more extreme end, \citet{longtail2023} remove all documents containing cooccurrences of terms from a subset of factual questions, ablating 30\% of the pretraining dataset in total before training a counterfactual model from scratch.

Different interventions come with their own pros and cons: token-level interventions enable studying more precise data-behavior relationships but may be too small to have an effect, while removing large sets of documents is more likely to alter behavior but introduces confounding factors like altered token frequencies, domain shift, etc. All of these considerations should be taken into account when designing one's own document intervention procedure.

In summary, instantiating our recipe requires choosing an item selection function $f_{\text{sel}}$, a document matching function $g_{\text{match}}$, and an intervention function $g_{\text{int}}$.

\section{Experimental Setup}
\label{sec:exp-setup}

We demonstrate our recipe through case studies where we intervene on pretraining documents to study their effect on knowledge acquisition. Our first study relates term cooccurrences to relational fact learning (\S\ref{sec:pararel-results}). Our second study also considers term cooccurrences, but relates them to a different phenomenon by studying gender bias in a coreference resolution task (\S\ref{sec:gender-bias-results}). In our final study, we experiment with multiple-choice question answering (MCQA) datasets MMLU, OpenBookQA, and SciQ to study knowledge acquisition while moving beyond the term cooccurrence setting (\S\ref{sec:mcqa-results}). We use a similar method of selecting items and general intervention procedure across studies, with specific variations discussed below and in the respective sections for each study. We use the same models and evaluation setup for all experiments.

\subsection{Model}

We use the OLMo models \cite{groeneveld2024, olmo2025} for our experiments, as they include all the necessary components for our recipe as listed in \S\ref{sec:components}.\footnote{The availability of such resources is essential for performing the kinds of analyses we describe, underscoring the importance of fully open LMs for studying pretraining.} Specifically, we use OLMo 2 1B and 7B with associated pretraining data, OLMo 2 Mix 1124 \cite{olmo2025}. For the 7B model we use the second and third checkpoints (at 2,000 and 3,000 steps, respectively). We create additional checkpoints at steps 2,150 and 3,150 (retraining over the original ordered data starting from steps 2,000 and 3,000, respectively) to perform interventions over smaller data batches. For the 1B model, we retrain the model from scratch for 10,000 steps on the original ordered data batches, checkpointing every 500 steps to yield more granular checkpoints.

\subsection{Evaluation}

Key components of our analysis require evaluating checkpoints on the items in each dataset, and we use OLMES to perform these evaluations \citep{olmes2025}. The OLMES framework outlines a variety of choices for prompting the model and ranking answer choices to standardize evaluations for MCQA tasks, which fits all of our datasets.

Following OLMES notation, we use the cloze/completion task formulation (CF) with a QA prompt as input (e.g., ``Question: The capital city of France is \textbackslash nAnswer:''). We use this format for all datasets and experiments. We rank LM probabilities of candidate answer choices with different normalizations (detailed in each section) based on which gives the best performance. We label items as correct if the correct response is ranked highest and calculate accuracy as our evaluation metric {$\mathcal{F}$}. We use zero-shot prompting for all datasets.

\subsection{Selecting Items}
\label{sec:learned-items}

For our factual knowledge studies, we perform two kinds of interventions for a particular data batch: (1) replacing documents that match items learned at a particular checkpoint with unrelated documents in order to \emph{suppress learning}; and (2) replacing unrelated documents with documents that match items learned at a later checkpoint to \emph{promote learning} earlier. To do so, we identify the first checkpoint at which certain facts (represented by items in our evaluation dataset) are ``definitively learned''. We define a checkpoint at which a fact is first ``definitively learned'' as the one where its item is first answered correctly (being incorrect at all previous checkpoints) and remains correct until a certain future checkpoint, specifically checkpoint 20 (10K steps) for OLMo 2 1B and checkpoint 30 (30K steps) for OLMo 2 7B in our experiments. Our definition enforces correctness for some future checkpoints to reduce the risk of targeting items labeled correctly by chance.

For our gender bias study, we perform a single type of intervention that aims to mitigate bias exhibited by the model, defined as a difference in task performance that favors one gender over another. In this setting, we simply select target items as those which exhibit bias at a particular checkpoint.

The early stages of pretraining are where the OLMo models exhibit the most improvement in performance for all datasets. We also find that the early checkpoints are where most items are learned (by our definition), with at least 80\% of learned items being labeled correctly by step 4,000 for the 1B model and step 3,000 for the 7B model (see Appendix Fig.~\ref{fig:learned-items}). For items learned at a particular checkpoint, we identify the subset that have some number of matching documents in the data batch immediately preceding that checkpoint and use those as our target subset $\mathcal{E}^{\text{target}}$.

\subsection{Intervening}
\label{sec:intervening}

For our factual knowledge experiments, we identify a pair of consecutive model checkpoints $\mathcal{M}_t$ and $\mathcal{M}_{t+1}$ where we either want to \emph{promote learning} of items learned at $\mathcal{M}_{t+1}$ (but not yet learned at $\mathcal{M}_t$) or \emph{suppress learning} of items learned at $\mathcal{M}_t$. In either setting, we intervene on the data batch $\mathcal{D}_t$ just prior to $\mathcal{M}_t$, with details on each intervention described separately for each experiment. In general, we attempt to \emph{suppress learning} by replacing matching documents in $\mathcal{D}_t$ with unrelated ones from $\mathcal{D}_{t+1}$ , and we \emph{promote learning} by moving matching documents from $\mathcal{D}_{t+1}$ into $\mathcal{D}_t$ (replacing unrelated documents to do so). Specific details on this document-swapping operation can be found in Appendix \S\ref{sec:doc-swapping}.

For our gender bias experiments, we similarly identify a pair of consecutive model checkpoints {$\mathcal{M}_t$} and {$\mathcal{M}_{t+1}$} where we want intervene on the data batch between them to modify some behavior observed at checkpoint {$\mathcal{M}_{t+1}$}. Rather than swapping whole documents, our intervention swaps the gender of pronouns in documents to alter term frequencies in a way that mitigates the observed bias (more details can be found in \S\ref{sec:gender-bias-results}).

After constructing our intervention data batch $\mathcal{D}_t^{\text{int}}$, we take $\mathcal{M}_{t-1}$ and continue training up to the next checkpoint using the same optimizer state, learning rate, and learning rate schedule as the original OLMo 2 pretraining run at that point. Other than the intervened documents, the model will see the same documents in the same order as the original model that was trained from $\mathcal{M}_{t-1}$ to $\mathcal{M}_t$, allowing us to construct an intervention checkpoint $\mathcal{M}_t^{\text{int}}$ that exhibits the effects of only our intervention on the original data batch $\mathcal{D}_t$.

Interestingly, we observe different results even with repeated runs of the exact same portion of pretraining (i.e., the same initial state and ordered data batches). We attribute this to irreducible noise arising from distributing training across GPUs. We repeat training runs over each intervention data batch as well as the original data batch to take this variation into account, reporting the mean and standard deviation across runs for each condition.
\section{Impact of Term Cooccurrences on Relation Fact Learning}
\label{sec:pararel-results}

In the first study, we test the hypothesis that the presence of term cooccurrences -- for example, the cooccurrence of ``Paris'' and ``France'' given the factual relation ``Paris is the capital of France'' -- in the data batch immediately prior to a fact being learned affects whether it is learned at all. Existing work has studied the relationship between term cooccurrence and factual knowledge acquisition either through observational data \citep{termfreq2022, longtail2023, measuringcausal2022, kang2023, zhang2024} or by pretraining from scratch on synthetic data \citep{kassner2020} or modified pretraining datasets \citep{wei2021, longtail2023}. In contrast to past work, which primarily analyzes the \emph{final} pretrained model, we study the intermediate checkpoints where these facts are \emph{first} learned, by making specific, targeted interventions on the data.

\subsection{Experiment Setup}

\textbf{Dataset} We use ParaRel\footnote{We use a version of ParaRel that was modified for autoregressive language modeling \citep{coastal2024}.} as our evaluation dataset $\mathcal{E}$ \citep{elazar2021}, which consists of WikiData facts formatted as a relation between two entities -- a subject and an object -- allowing us to use subject-object cooccurrence as a heuristic to identify matching documents. Each relation is also associated with various templates that express it in natural language, and we use a single template per relation when querying our LM. We use the 12 relations where both models show better than chance performance at the intervention checkpoint, with full details on those relations in Appendix Table~\ref{tab:pararel-details}. When evaluating, we use no length normalization when ranking candidate answer choices, after comparing to all other OLMES normalization strategies and finding that no normalization gave the best performance for this dataset.

\textbf{Counting (co)occurrences} We count occurrences of entities from ParaRel items mentioned in a batch of pretraining data using WIMBD\footnote{Specifically, we use the \texttt{wimbd search} command for fast exact string matching.} \citep{elazar2024}. This allows us to identify the documents entities occur in and their starting and ending character location within those documents. For items where the strings of the subject and object have partial overlap (e.g., ``Microsoft Windows is a product of Microsoft''), we only consider cooccurrences where the character windows of the two terms do not overlap. We note that cooccurrence is a heuristic which does not guarantee that each matched document is truly relevant. In Appendix~\ref{sec:doc-relevance}, we analyze the relevance of matched documents and show that only a fraction of those that contain a cooccurrence can be used to correctly answer the factual knowledge question, although a majority are still broadly relevant.

\textbf{Intervening} We attempt to suppress and promote learning of items originally learned at two different pretraining steps each for the 1B and 7B OLMo 2 models, retraining over steps 1.5--2K and 2.5--3K for the 1B model and steps 2--2.15K and 3--3.15K for the 7B model. In each case, we attempt to suppress learning of items originally learned at the end of each data batch, and promote learning of items learned at the end of the next batch. To suppress learning, we use both cooccurrence of subject and object entities as well as occurrence of either entity to identify related documents. The replacement of occurrences is a more aggressive intervention to determine a lower bound on performance change. This results in replacing 0.5--4.7\% of documents for cooccurrences and 17.3--43.7\% for occurrences, depending on the model size and data batch. For promoting learning, we only use cooccurrences to identify relevant documents to move into the previous batch, replacing 1.3--9.9\% of documents for any given model size and pretraining step.

In all cases, we initialize from the OLMo 2 checkpoint at the initial step and continue pretraining for 500 (1B) or 150 (7B) steps on the modified data batch, then evaluate the counterfactual checkpoints on  our target subset for which we are attempting to suppress or promote learning. We perform 5 repeated training runs for each result. Additional details on each intervention can be found in Appendix Table~\ref{tab:pararel-intervention-stats}.

\subsection{Result: Removing Cooccurrences and Occurrences Suppresses Learning}

\begin{figure*}[!t]
    \centering
    \includegraphics[width=0.8\textwidth]{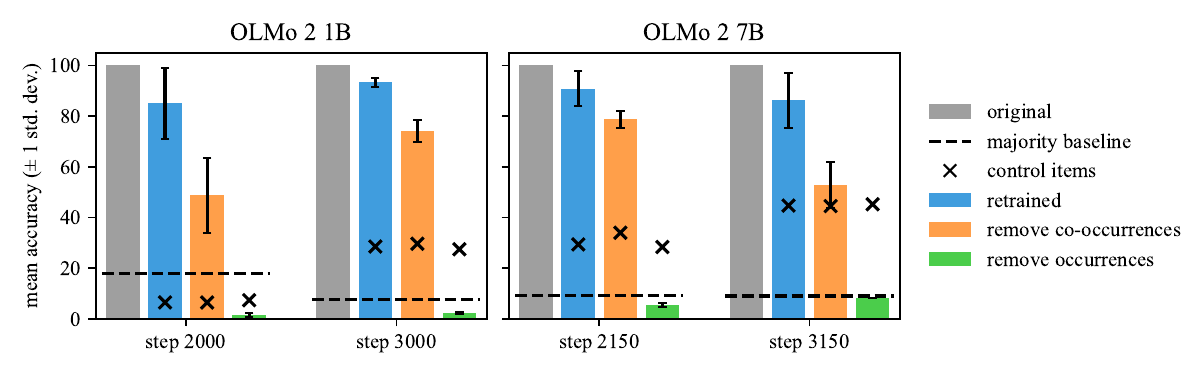}
    \caption{Results for replacing matching documents and re-training checkpoints to \textit{suppress learning} of selected items from ParaRel. Colored bars show accuracy on the subset of target items for each intervention condition, including retraining on the original data batch (\textcolor{tab:blue}{blue}), training on a modified batch that replaces cooccurrences (\textcolor{tab:orange}{orange}), and training on a modified batch that (more aggressively) replaces occurrences (\textcolor{tab:green}{green}). We also show results on control items ({$\mathbf{\times}$}; i.e., all items other than the target group). For all results, we perform 5 repeated training runs and show the mean and one standard deviation across runs.}
    \vspace{-.4cm}
    \label{fig:suppress-learning}
\end{figure*}

Fig.~\ref{fig:suppress-learning} shows that performance for learned items is reduced when cooccurrences are removed, and reduced even further when removing occurrences. This trend holds for multiple checkpoints and across model scales. We also compare all results to a \emph{majority baseline} that  selects the most commonly-occurring object entity for each relation in the ParaRel training set.

Notably, removing all cooccurrences yields higher performance than the majority baseline, with a 30.7--66.5\% absolute difference in accuracy depending on the model size and pretraining step. If an LM's transition from answering an item incorrectly at one checkpoint to correctly at the next was entirely due to cooccurrences in the data seen between those two checkpoints, we would expect removing those cooccurrences to drop performance to the majority baseline. As this is not the case, we conclude that while cooccurrences in the most recent data batch encourage learning, they are not responsible for all of it. The remaining difference could be explained by a number of factors -- cooccurrences from previous data batches, multi-hop cross-document chains of relations linking the two entities, cooccurrences of different terms with similar model representations, etc. -- all of which would be interesting future interventions for further understanding this phenomenon.

In contrast, the more drastic intervention of removing occurrences is sufficient to completely suppress learning for all steps and model scales, although this modifies a much larger fraction of each data batch. This suggests that learning is in fact caused by the documents with entity mentions, although not just by documents that contain cooccurrences. Testing other hypotheses could determine whether the same reduction in performance could be obtained by a smaller intervention.

Finally, we also show the effect of our intervention on a \emph{control group} of all items other than those we target for suppressing learning. Change in control group performance is minimal, indicating that our intervention successfully targets items for which we aim to alter performance. This strengthens the claim that the documents we intervene on have a direct relationship with the target evaluation items, as opposed to having a more general effect on task performance.

\subsection{Result: Introducing Cooccurrences Promotes Learning}

\begin{figure}
    \centering
    \includegraphics[width=0.5\textwidth]{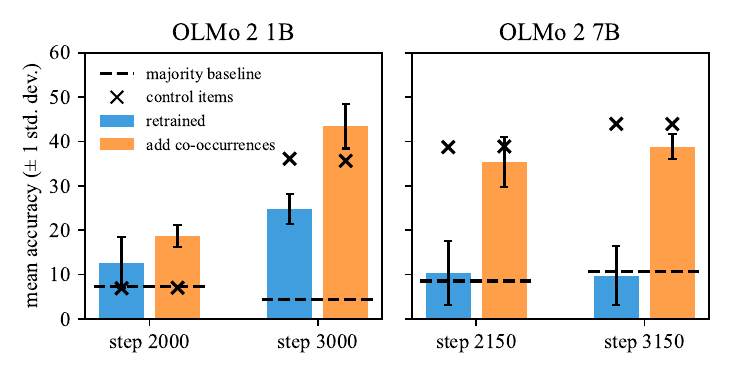}
    \caption{Results for replacing matching documents and retraining to \emph{promote learning}. Replacing documents not containing a cooccurrence with documents that do improves performance in all cases.}
    \vspace{-.4cm}
    \label{fig:promote-learning}
\end{figure}

Fig.~\ref{fig:promote-learning} shows that introducing cooccurrences promotes learning, increasing performance on the target items above their original performance and well above the majority baseline. However, these items were selected because they were all labeled correctly (i.e., ``learned'') at the very next checkpoint, and our intervention fails to attain perfect accuracy. As in the previous section, this finding suggests that documents beyond the ones that contain entity cooccurrences in this data batch (or possibly a prior one) also contribute to the model's acquisition of factual knowledge for these items. As in the learning-suppression setting, our intervention has minimal impact on control group items.

We perform supplementary experiments varying our method of matching documents to items (Appendix Fig.~\ref{fig:vary-matching}) and our method of intervening on matched documents (Appendix Fig.~\ref{fig:vary-modifying}). These experiments show a similar pattern to what is reported above.

\begin{table*}[t!]
    \centering \small
    \setlength{\tabcolsep}{8pt}
        \begin{tabular}{@{} l l c c c c c c @{}}
        \toprule
         & & \multicolumn{2}{c}{MMLU} & \multicolumn{2}{c}{OpenBookQA} & \multicolumn{2}{c}{SciQ} \\
        \cmidrule(lr){3-4}\cmidrule(lr){5-6}\cmidrule(lr){7-8}
        Model & Intervention & acc. & $\Delta$acc. & acc. & $\Delta$acc. & acc. & $\Delta$acc. \\
        \midrule
        \multirow{4}{*}{OLMo 2 1B}
         & retrained                           & $89.8_{0.8}$ & --      & $88.7_{2.0}$ & --      & $89.8_{1.4}$ & -- \\
         & replace high-scoring BM25 docs      & $76.0_{0.6}$ & $-13.8$ & $70.2_{3.0}$ & $-18.5$ & $48.3_{2.3}$ & $-41.5$ \\
         & replace high-scoring DPR docs       & $86.0_{2.8}$ & $-3.8$  & $78.0_{2.0}$ & $-10.7$ & $59.4_{5.7}$ & $-30.4$ \\
         & majority baseline                   & $26.9$       & $-62.9$ & $27.6$       & $-61.1$ & $25.0$       & $-64.8$ \\
        \midrule
        \multirow{4}{*}{OLMo 2 7B}
         & retrained                           & $84.8_{3.7}$ & --      & $78.0_{4.9}$ & --      & $80.6_{2.5}$ & -- \\
         & replace high-scoring BM25 docs      & $46.8_{2.1}$ & $-38.0$ & $65.3_{4.1}$ & $-12.7$ & $50.6_{2.4}$ & $-30.0$ \\
         & replace high-scoring DPR docs       & $57.1_{6.5}$ & $-27.7$ & $70.0_{1.6}$ & $-8.0$  & $55.0_{1.6}$ & $-25.6$ \\
         & majority baseline                   & $26.9$       & $-57.9$ & $27.6$       & $-50.4$ & $25.0$       & $-55.6$ \\
        \bottomrule
        \end{tabular}
    \caption{Results on \emph{suppressing learning} for MCQA datasets. Results show the mean and standard deviation (subscript) across 3 training runs. $\Delta$acc.\ is the change in accuracy as compared to the mean accuracy of the retrained model.}
    \label{tab:mcqa-suppress}
\end{table*}

\section{Impact of Pronoun-Profession Cooccurrence on Gender Bias}
\label{sec:gender-bias-results}

Our recipe is intended to be general enough to analyze a broad range of phenomena. To test generality, we perform experiments altering cooccurrence of gendered pronouns and professions to affect gender bias behavior.

\subsection{Experiment Setup}

\noindent\textbf{Dataset} We use the WinoBias coreference resolution dataset \cite{zhao-etal-2018-gender}, which contains 783 sentence pairs, where each sentence mentions two professions and a gendered pronoun (e.g., ``\emph{The developer} argued with \emph{the designer} because \emph{she} did not like the design.'') and sentences in each pair differ only in the pronoun's gender. For evaluation, we prompt the model with a sentence and a question asking which profession the pronoun refers to, and select the answer by ranking each profession by its log-probability. We find that all OLMES length normalization strategies perform comparably, so we use no length normalization.

\noindent\textbf{Quantifying bias} We measure \emph{gender bias} as the performance discrepancy when using a male versus a female pronoun with a profession. Specifically, we compute the accuracy difference between ``pro''-stereotype items (e.g., a majority-male profession with a male pronoun; majority determined using U.S. Bureau of Labor Statistics data) and ``anti''-stereotype items.

\noindent\textbf{Counting cooccurrences} Similar to ParaRel, we use regular expressions to match mentions of professions and pronouns. We do not enforce that pronouns and professions refer to the same entity, merely that they occur in the same document.

\noindent\textbf{Selecting items} Selected items are sentence pairs where the model is correct for the ``pro''-stereotype profession-pronoun sentence and is incorrect for the ``anti''-stereotype one, showing bias in favor of the ``pro''-stereotype gender-profession pair. We perform separate experiments for target items where the ``pro''-stereotype is male (i.e., majority-male professions) and those where it is female, as we observe bias favoring each gender depending on the profession.

\noindent\textbf{Intervening} We intervene on documents that contain a profession for which the model shows bias by changing all ``pro''-stereotype pronouns to ``anti''-stereotype ones. This gives two separate interventions for different professions, swapping all pronouns to either female or male. To preserve grammaticality, we resolve the ambiguous case of possessive pronouns (e.g., ``his book'' {$\rightarrow$} ``her book'' vs. ``the book was his'' {$\rightarrow$} ``the book was hers'') by using the spaCy dependency parser \cite{spaCy2020} to identify the dependency label of each pronoun. We intervene between steps 5.5--6K for OLMo 2 1B and steps 2--2.15K for OLMo 2 7B, and perform 3 repeated runs. Additional details on each intervention can be found in Appendix Table~\ref{tab:gender-bias-intervention-stats}.

\subsection{Result: Swapping Gendered Pronouns Mitigates Gender Bias}

\begin{figure}
    \centering
    \includegraphics[width=0.5\textwidth]{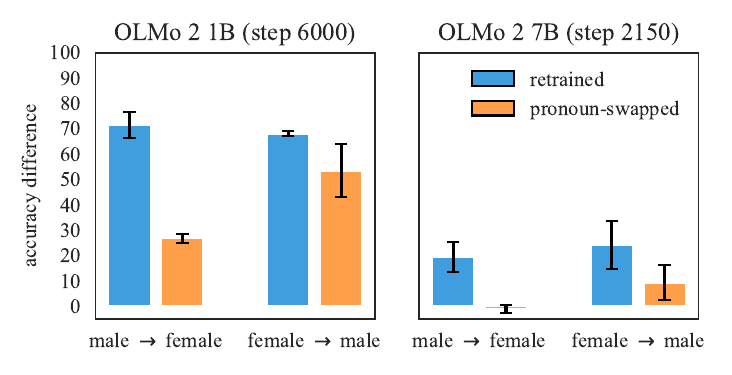}
    \caption{Results for swapping the gender of pronouns in documents that mention stereotyped professions to mitigate gender bias. Swapping pronouns is an effective intervention for mitigating bias across model sizes and stereotyped profession subsets (male and female).}
    \vspace{-.4cm}
    \label{fig:gender-bias}
\end{figure}

Results of our interventions are shown in Fig.~\ref{fig:gender-bias}. Swapping the gender of pronouns in documents mentioning stereotyped professions mitigates gender bias across both model sizes and both profession subsets (male- and female-stereotyped), as reflected by a decrease in the accuracy difference between ``pro''- and ``anti''-stereotyped sentences. The 1B model exhibits greater overall bias and a more pronounced reduction under the intervention. These results corroborate \citet{biderman2023}, showing that this intervention is effective for a newer model family and at intermediate stages of pretraining. They also complement our findings on term cooccurrence and knowledge acquisition, and demonstrate that our recipe applies to a broad range of data-behavior relationships.

\section{Impact of Retrieved Documents on Complex Knowledge Queries}
\label{sec:mcqa-results}

In \S\ref{sec:pararel-results}, we studied the relationship between pretraining data and factual knowledge for facts that can be easily expressed as a relation between a pair of entities. While this format provides a useful way of studying knowledge acquisition via term cooccurrence, this is not always the way knowledge is expressed. Modern MCQA datasets test LM knowledge through complex questions paired with several possible answers, with questions and answers sometimes containing references to multiple entities or expressing abstract concepts involving no salient entities at all. Since term cooccurrence cannot be easily used to identify relevant documents for these items, we instead use methods from information retrieval (IR). As with our experiments on ParaRel, we believe that information in pretraining documents is contributing to knowledge learning for our MCQA datasets, and we hypothesize that IR methods are an effective way of identifying documents that contribute to knowledge acquisition.

\subsection{Experiment Setup}

\textbf{Datasets} We use common datasets for evaluating or comparing models \cite{groeneveld2024, olmo2025, olmes2025}: MMLU \cite{hendrycks2021measuring}, OpenBookQA \cite{mihaylov-etal-2018-suit}, and SciQ \cite{welbl-etal-2017-crowdsourcing}. OpenBookQA and SciQ consist of exam-style science questions, while MMLU contains questions from 57 subjects. For all datasets, each item includes a question with 4 multiple-choice answer options which we rank according to the cloze formulation. We follow OLMES when performing length normalization for these datasets, using PMI normalization for OpenBookQA and character normalization for MMLU and SciQ (full details in Appendix Table~\ref{tab:mcqa}).

\textbf{Retrieval-based matching} Since we often cannot easily identify terms from items in these datasets to search for in pretraining documents, we instead use IR methods to match items to documents. Specifically, we use the keyword-based method BM25 \cite{roberts-etal-2020-much} and the embedding-based method Dense Passage Retriever (DPR; \citealp{karpukhin2020}). BM25 is a popular method that scores a document's relevance to a search query based on relative word frequencies,\footnote{We use the \texttt{bm25s} Python package.} while DPR uses trained models to compute item and document embeddings and scores a document's relevance to an item using embedding similarity.\footnote{We use the multiset DPR encoders from Facebook \cite{wolf-etal-2020-transformers}.} We include both the question and the correct answer in the search queries for BM25 and to construct query embeddings using DPR. As in \S\ref{sec:pararel-results}, we include additional analysis of document relevance in Appendix~\ref{sec:doc-relevance}.

\textbf{Intervening} For our MCQA datasets, we focus on the intervention to \emph{suppress learning}. When using BM25 and DPR to identify matching documents, we use the fact that each method assigns a real value similarity score to each document-item pair to construct our intervention. Specifically, we replace the top-$k$ highest-scoring documents from $\mathcal{D}_t$ the lowest-scoring documents from $\mathcal{D}_{t+1}$, while matching token counts. We match documents greedily, iterating over relevant documents in order of greatest to least number of matching items and finding a document to replace each one. In practice, this results in a lower average relevance score across replaced documents.

We intervene at steps 2K--2.5K for OpenBookQA and SciQ and steps 1K--1.5K for MMLU at the 1B scale, and steps 2K--2.15K for all datasets at the 7B scale; these steps were chosen because they correspond to stages when many items were learned. In all cases, we replace $k$ = 1K documents per item,\footnote{Exploratory experiments suggested that increasing to 2K documents per item changed the results very little.} which results in replacing 9.3--11.1\% of documents for MMLU, 3.5--6.6\% of documents for OpenBookQA, and 10.5-21.3\% of documents for SciQ, depending on the method and model scale. We perform 3 repeated training runs for each setting. Additional details on each intervention can be found in Appendix Table~\ref{tab:mcqa-intervention-stats}.

\subsection{Result: Removing Relevant Documents Suppresses Learning}

Results are in Table~\ref{tab:mcqa-suppress}, with all interventions leading to a decrease in average accuracy. Using DPR scores seems to have less of an effect than using BM25 scores across datasets and model scales, even when matching the number of replaced documents per item. Similar to ParaRel, the performance after our intervention is still well above the baseline for all datasets, suggesting that the documents identified by the IR methods are still not entirely responsible for knowledge learning.

We include additional results on \emph{promoting learning} for MCQA datasets in Appendix~\ref{app:mcqa-promote}, showing that adding relevant documents similarly increases accuracy on target group items.
\section{Related Work}

The most closely related work was highlighted earlier; here we note  additional strands of interest.

\paragraph{Training data attribution} Our work is  related to quantifying the impact of a training example on a model's prediction \cite{Hammoudeh2022TrainingDI}. Among the most popular methods for doing so are gradient-based techniques such as influence functions \cite{koh2017}, which have been applied to LM fine-tuning data \cite{han2020} as well as pretraining data \cite{grosse2023}, with the latter requiring approximations due to high computational cost. These approaches have their own limitations, as they sometimes perform comparably to cheaper alternatives like BM25 at identifying relevant documents \cite{akyurek-etal-2022-towards} and may assign high influence to documents which do not express the information \cite{chang2025scalable}. While often used to explain a model's predictions rather than test hypotheses about how data causes behavior, influence functions could be integrated into our recipe for identifying target documents to intervene on.

\paragraph{Continued training} In contrast to our method of intervening at intermediate steps, many other works continue training from the final checkpoint on select data to study behavior or improve LMs. \citet{razeghi2023backtracking} continue pretraining on documents selected via embedding similarity or regular expression matching to improve math reasoning. Embedding or gradient similarity has also been used to select data for continued training to adapt LMs to domains and tasks \cite{gururangan-etal-2020-dont}, boost performance on certain tasks \cite{han2022}, improve in-context learning ability \cite{han-etal-2023-understanding}, or improve instruction-following \cite{ivison-etal-2023-data, xia2024}. Notably, the last set of approaches focus on boosting performance rather than understanding LM behavior.
\section{Discussion and Conclusions}

We have introduced a flexible recipe for interventional studies of the effects of pretraining data on LM behavior, formalizing each stage to enable systematic exploration. Our recipe highlights the importance of targeted interventions built around specific test items and studying changes at intermediate steps, not merely the final model.

Our results show that simple interventions that add or remove documents relevant to the fact (either using term cooccurrence or information retrieval), have a consistent effect on promoting and supressing learning of facts. These findings support the position that exposure to a textual expression of a fact in training helps a model learn to correctly report it at test time. However, adding relevant documents does not guarantee learning, and removing them does not always prevent learning. More study is required to fully characterize the process by which language models acquire facts, and we believe the methodological tools given by our recipe will aid those future efforts.
\section{Limitations}

As we designed our recipe to be general enough to test a variety of hypotheses, we focus here on the limitations of our case studies on knowledge acquisition in LMs.

Our interventions are highly localized, retraining between early checkpoints while modifying a small portion of the pretraining dataset due to computational constraints. While this limits the extent to which our findings apply to the full pretrained model, localized interventions can still yield insights for practitioners and model builders into how models respond to changes in pretraining data, as shown in our case studies as well as in previous smaller-scale studies \cite{biderman2023, howdollms2024}. Importantly, our recipe could be applied to larger portions of pretraining or even a full pretraining run given more computational resources, providing a methodological contribution beyond localized settings.

In our experiments, the fact that our interventions are only partially effective may be due to documents from past data batches that we did not intervene on, and a more involved study could apply our recipe to test this hypothesis. The effects of our interventions may also be weakened by other factors. While cooccurrence and information retrieval are well-established methods of identifying potentially relevant documents, they are based on heuristics, and may not identify \emph{all} documents that express a particular fact (see Appendix~\ref{sec:doc-relevance} for further analysis). LMs may also make connections across multiple documents, such that matching individual documents is insufficient. Future work could expand the notion of a ``document'' to address this challenge when performing matching.

\section*{Acknowledgments}
We thank the anonymous reviewers and the action editor, André F. T. Martins, for their valuable suggestions. This research is supported by the National Artificial Intelligence Research Resource (NAIRR) Pilot and the Anvil supercomputer supported by the National Science Foundation (award NSF-OAC 2005632) at Purdue University, as well as NVIDIA resources provided through NAIRR.

\bibliography{tacl2021v1}
\bibliographystyle{acl_natbib}

\appendix
\section{Appendix}

\subsection{Swapping Documents Between Batches}
\label{sec:doc-swapping}

\begin{figure}[h!]
    \centering
    \includegraphics[width=0.5\textwidth]{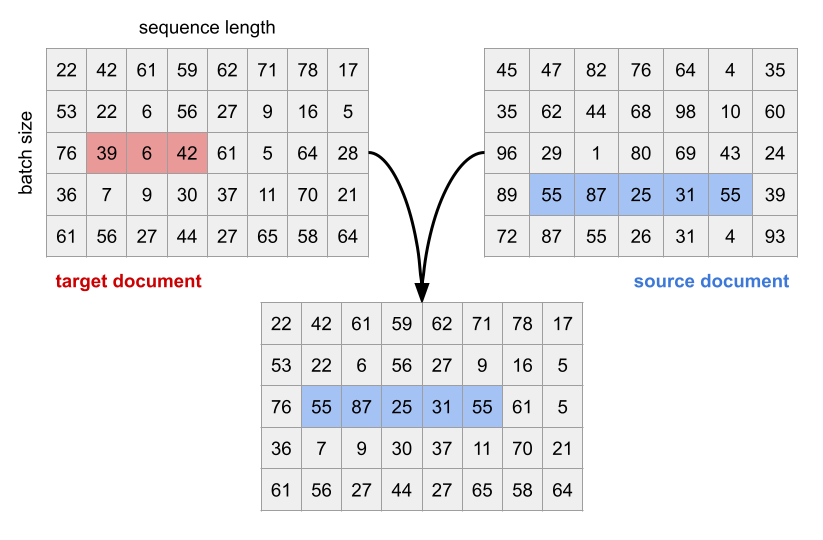}
    \caption{Toy example that shows how token IDs in a batch of tokens are altered when replacing a target document with a source document. Note how the token IDs to the right of the target document are shifted to the right to accommodate the longer source document.}
    \label{fig:replace-tokens}
\end{figure}

All of our interventions involve taking documents from another data batch and using them to replace documents in the data batch that we wish to intervene on. We replace documents while matching total token counts when possible to minimize altering the data batch. For over 96\% of documents, the document that replaces it has exactly the same number of tokens. In a few cases, the document being swapped in is too long by a few tokens, which requires shifting the remaining tokens in the sequence to the right and truncating at the end. Fig.~\ref{fig:replace-tokens} shows a toy example of how this document-swapping operation is performed with token IDs across two batches of tokens.

\subsection{Result: Adding Relevant Documents Promotes Learning}
\label{app:mcqa-promote}

\begin{table*}[t!]
    \centering \small
    \setlength{\tabcolsep}{8pt}
        \begin{tabular}{@{} l l c c c c c c @{}}
        \toprule
         & & \multicolumn{2}{c}{MMLU} & \multicolumn{2}{c}{OpenBookQA} & \multicolumn{2}{c}{SciQ} \\
        \cmidrule(lr){3-4}\cmidrule(lr){5-6}\cmidrule(lr){7-8}
        Model & Intervention & acc. & $\Delta$acc. & acc. & $\Delta$acc. & acc. & $\Delta$acc. \\
        \midrule
        \multirow{4}{*}{OLMo 2 1B}
         & retrained                           & $21.3_{0.0}$ & --     & $15.2_{5.4}$ & --      & $20.4_{3.3}$ & --      \\
         & replace high-scoring BM25 docs      & $29.1_{4.8}$ & $+7.8$ & $22.7_{0.0}$ & $+7.5$  & $47.1_{0.7}$ & $+26.7$ \\
         & replace high-scoring DPR docs       & $22.0_{1.3}$ & $+0.7$ & $13.6_{2.3}$ & $-1.6$  & $34.3_{7.2}$ & $+13.9$ \\
         & majority baseline                   & $26.9$       & $+5.6$ & $27.6$       & $+12.4$ & $25.0$       & $+4.6$  \\
        \midrule
        \multirow{4}{*}{OLMo 2 7B}
         & retrained                           & $13.6_{3.6}$ & --      & $14.9_{3.0}$ & --      & $11.9_{3.7}$ & --      \\
         & replace high-scoring BM25 docs      & $53.7_{4.6}$ & $+40.1$ & $34.5_{7.0}$ & $+19.6$ & $59.4_{2.0}$ & $+47.5$ \\
         & replace high-scoring DPR docs       & $31.9_{2.6}$ & $+18.3$ & $25.3_{5.2}$ & $+10.4$ & $49.6_{1.1}$ & $+37.7$ \\
         & majority baseline                   & $26.9$       & $+13.3$ & $27.6$       & $+12.7$ & $25.0$       & $+13.1$ \\
        \bottomrule
        \end{tabular}
    \caption{Results on \emph{promoting learning} for MCQA datasets. Results show the mean and standard deviation (subscript) across 3 training runs. $\Delta$acc. is the change in accuracy as compared to the mean accuracy of the retrained model.}
    \vspace{-.4cm}
    \label{tab:mcqa-promote}
\end{table*}

Table~\ref{tab:mcqa-promote} shows results for \emph{promoting learning}, where similarly to suppressing learning all of our interventions cause a change in average accuracy. The change in performance is again larger when using BM25 scores rather than DPR scores across all datasets and model scales. In this setting, there is a consistently larger absolute increase in accuracy at the 7B model scale for all datasets, a pattern which was not present when suppressing learning. Given existing results showing that performance improves more for larger models with methods like instruction fine-tuning \citep{JMLR:v25:23-0870} or parameter-efficient techniques like prompt-tuning \citep{lester-etal-2021-power}, this effect may be the consequence of larger models being more amenable to performance improvement.

\subsection{Determining Document Relevance}
\label{sec:doc-relevance}

While cooccurrence, BM25, and DPR are all reasonable methods of matching documents to items, ultimately they are approximations of our primary objective: identifying documents that express the information of a particular item. We assess the relevance of the documents matched by each of these methods by using GPT-5 \citep{openai_gpt5} as a judge to annotate a subset of documents. While using GPT-5 to perform the actual document matching over an entire data batch containing millions of documents is not scalable, using GPT-5 to annotate a subset of documents allows us to analyze the performance of our other more efficient matching methods. We also perform manual annotation on a smaller subset to compare to the GPT-5 annotations.

\begin{figure*}[t!]
    \centering
    \includegraphics[width=\textwidth]{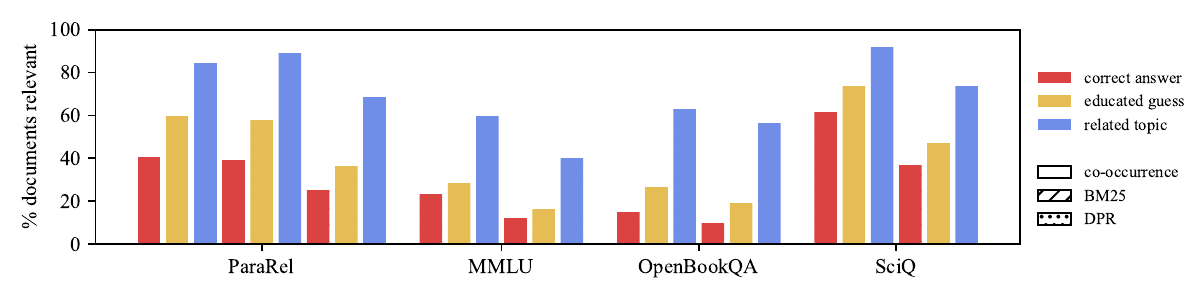}
    \caption{Percentage of documents labeled relevant to their matching item by \textbf{GPT-5}, for every dataset, prompt, and matching method.}
    \label{fig:gpt5-annotations}
\end{figure*}

\begin{figure*}[t!]
    \centering
    \includegraphics[width=\textwidth]{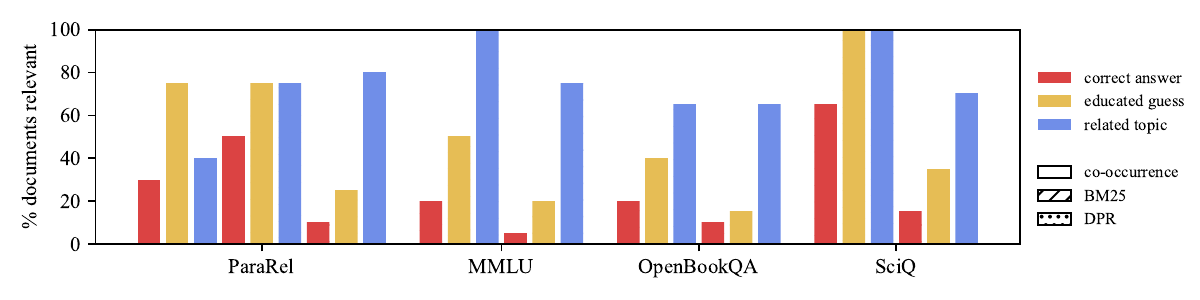}
    \caption{Percentage of documents labeled relevant to their matching item by \textbf{manual annotation}, for every dataset, prompt, and matching method.}
    \label{fig:manual-annotations}
\end{figure*}

\paragraph{Setup}

For each combination of dataset (ParaRel, MMLU, OpenBookQA, and SciQ) and matching method (cooccurrence (ParaRel only), BM25, and DPR), we select the top-20 highest-scoring documents for each intervention group item from our OLMo 2 1B experiments to form a subset for annotation.\footnote{Since cooccurrence does not provide a score for each document, we instead use a random sample of 20 matching documents.}

For each item-document pair, we provide GPT-5 (\texttt{gpt-5} in the OpenAI API, accessed September 28, 2025) with the question, correct answer, and document along with one of three prompts to elicit ``yes''/``no'' annotation of varying levels of document relevance. We use ``correct answer'', ``educated guess'', and ``related topic'' prompts, which are designed to determine whether the document contains sufficient information to correctly answer the question, contains enough information to make an educated guess about the answer (while not containing the answer), or is topically related to the question despite not containing enough information to answer it, respectively. The specific format of each prompt can be found in Table~\ref{tab:gpt5-prompts}. We set the verbosity to ``low'' and the reasoning level to ``minimal'' when querying the OpenAI API.

\paragraph{Results}

Table~\ref{tab:gpt5-annotation-examples} shows example items from MMLU with documents matched by BM25 score that were annotated by GPT-5 as being relevant or not, for each of the three annotation prompts. A summary of the GPT-5 annotation results can be found in Fig.~\ref{fig:gpt5-annotations}, where we plot the percentage of item-document pairs for which GPT-5 responded ``yes'' when asked to judge document relevance given each prompt. Overall, 9.8--61.3\% of documents are determined to be relevant according to the ``correct answer'' prompt, 15.9--59.6\% according to the ``educated guess'' prompt, and 39.8--91.8\% according to the ``relevant topic'' prompt, with percentages varying considerably across datasets. This suggests that while our matching methods can identify documents that are generally relevant to a question, they are less precise when it comes to identifying documents that directly entail the answer and could provide a model with enough information during pretraining to make a correct prediction.

For ParaRel, cooccurrence performs comparably to BM25 at identifying relevant documents across prompts, likely because each item has a simple format with the subject and object entity names being the most relevant keywords. DPR performs worse than BM25 across all datasets at identifying relevant documents. These results align with our interventional analysis findings, where using BM25 scores to select documents led to both greater promotion and suppression of learning (depending on the experiment) than using DPR scores.

\paragraph{Human validation}

To manually validate the GPT-5 annotations, we further select a random subset of 20 item-document pairs for each \{dataset, matching method\} pair which were then annotated by one of the authors following the same prompts. The corresponding manual annotation results can be found in Fig.~\ref{fig:manual-annotations}. Generally, the trends are similar to the GPT-5 annotations, although the exact percentages are different. We assess human/GPT-5 annotation agreement by computing Cohen's {$\kappa$} which gives a value of 0.54 overall, indicating a moderate level of agreement.

\begin{table*}[t!]
    \centering
    \resizebox{\textwidth}{!}{
    \begin{tabular}{l|p{5in}}
        \toprule
        \textbf{Prompt Name} & \textbf{Template} \\
        \midrule
        \multirow{5}{*}{correct answer} &
        You will be given a question or question stem and its correct answer, along with a document. Output ``yes'' if the document contains enough information to correctly answer the question, and ``no'' otherwise. Only output ``yes'' or ``no''. \\[4pt]
        & Question: \{question\} \\[2pt]
        & Answer: \{answer\} \\[2pt]
        & Document: \{document\} \\[4pt]
        \midrule
        \multirow{5}{*}{educated guess} &
        You will be given a question or question stem and its correct answer, along with a document. Output ``yes'' if the document contains enough information to make an educated guess about the answer, and ``no'' otherwise. The document does not have to have enough information to be able to correctly answer the question. Only output ``yes'' or ``no''. \\[4pt]
        & Question: \{question\} \\[2pt]
        & Answer: \{answer\} \\[2pt]
        & Document: \{document\} \\[4pt]
        \midrule
        \multirow{5}{*}{related topic} &
        You will be given a question or question stem and its correct answer, along with a document. Output ``yes'' if the document is topically related to the question and answer, and ``no'' otherwise. The document does not have to have enough information to be able to correctly answer the question. Only output ``yes'' or ``no''. \\[4pt]
        & Question: \{question\} \\[2pt]
        & Answer: \{answer\} \\[2pt]
        & Document: \{document\} \\[4pt]
        \bottomrule
    \end{tabular}
    }
    \caption{Prompt templates used for document relevance annotation with GPT-5.}
    \label{tab:gpt5-prompts}
\end{table*}

\begin{table*}[t!]
    \centering
    \resizebox{\textwidth}{!}{
    \begin{tabular}{p{2in}|l|p{2.5in}|p{2.5in}}
        \toprule
        \textbf{Item} & \textbf{Prompt} & \textbf{Document with ``yes'' annotation} & \textbf{Document with ``no'' annotation} \\
        \midrule
        Question: What do herpetologists study? Answer: Reptiles and amphibians & correct answer & What Classes Do I Need to Take in High School To Prepare for Getting a Zoology Degree? Zoologists study the impact of the environment on animal habitats... Some zoologists will specialize in a species, such as herpetologists, who study reptiles and snakes. Taking science courses while in high school will help you determine in which area of zoology you will specialize... This career may be physically demanding when spending long periods outdoors in rough conditions while researching. & Amphibians vs Reptiles Reptile? Or amphibian? Telling the difference. For most people, these animals are pretty much the same.... Definition Amphibian means ``living double lives'' on water and land. Reptile means ``creeping stealthily under cover of darkness.''... Tetrafauna AquaSafe – The complete water conditioner. \\
        \midrule
        Question: The use of 'indentured labour' in the nineteenth century involved: Answer: people being transported to the British colonies and forced to work for one employer under poor conditions & educated guess & History of Guyana Guyana history The recorded history of Guyana can be dated back to 1499, when Alonso de Ojeda's first expedition arrived from Spain at the Essequibo River... Subsequently, in 1746, the Dutch authorities opened the area near the Demerara River to British immigrants because they were eager to attract more settlers... By 1786 the internal affairs of this Dutch colony were effectively under British control, though two-thirds of the plantation owners were still Dutch.  Africans and Indians Africans were enslaved and transported to Guyana as slaves; however, East Indians came as indentured labourers who worked in order to provide for their families back home. By the 1660s, the enslaved & Banner Viadrina Keynote Lecture: Prof. Dr. Andreas Eckert  Blurred Frontiers. ``Unfree'' and ``Free'' Labour since 1800 in a global perspective In the world of work relations and labour policies, the distinction between ``free'' and ``unfree'' labour has been crucial... This lecture critically discusses widespread assumptions which link ``unfree'' or forced labour to pre-industrial societies and instead argues that there was a close and complex connection between definitions and practices of unfree labor in Europe and in the colonies... The talk will present various historiographical debates related to the issue of unfree/free labour, but will also focus on what historical actors in different parts of the world themselves understood by unfree/free labour. \\
        \midrule
        Question: When you squeeze an air-filled party balloon, you increase its Answer: density & related topic & hydrogen is 0.09 kg/m3. Two balloons are inflated to the same size, one filled with air and the other filled with hydrogen. Which balloon experiences the greatest buoyant force? 6) A ten kilogram brick is lowered into a tub of water. A volume equal to two litres of water is displaced. If the density of water is 1g/cm3 what is the reading of the scale when the brick is fully submerged? Does the brick lose mass when it is placed in water? Activity to measure the buoyant force. Put-Put boats and buoyancy & The Quartet Machine My instrument is called ``The Quartet Machine.'' It is a member of the string and woodwind families.  It is made out of a wooden box, rope, string, beads, plastic bar, rubber bands, balloon, barrette, ``squeakies,'' and a pencil... The second is you can blow air into the balloon and pinch it near the opening and pull. The third is you can squeeze the ``squeakies.''  To change the pitch of this instrument you can do one of the following...  Finally, to change the dynamics or volume of this instrument you can pluck the strings harder, squeeze the balloon harder or pinch the ``squeakies'' tighter. \\
        \bottomrule
    \end{tabular}
    }
    \caption{A sample of items from MMLU with example documents matched by BM25 score that were given ``yes'' and ``no'' annotations by GPT-5, for each of the three annotation prompts.}
    \label{tab:gpt5-annotation-examples}
\end{table*}

\begin{table*}[t!]
    \centering
    \resizebox{\textwidth}{!}{
    \begin{tabular}{l|c|l|l}
        \toprule
        Relation & \# Items & Template & Example \\
        \midrule
        P17 & 928 & [X] is located in [Y] & Mezhdurechensky District is located in Russia \\
        P27 & 966 & [X] is a citizen of [Y] & Marc Lavoie is a citizen of Canada \\
        P36 & 681 & The capital city of [X] is [Y] & The capital city of Greece is Athens \\
        P127 & 687 & [X] is owned by [Y] & Colchester Community Stadium is owned by Colchester \\
        P131 & 881 & [X] can be found in [Y] & Ryerson University can be found in Ontario \\
        P138 & 604 & [X] is named after [Y] & Great Britain is named after Britannia \\
        P176 & 961 & [X] is a product of [Y] & Ferrari 365 is a product of Ferrari \\
        P178 & 580 & [X] is a product of [Y] & Windows Embedded Compact is a product of Microsoft \\
        P276 & 877 & [X] is located in [Y] & 1996 Summer Olympics is located in Atlanta \\
        P495 & 907 & [X] was created in [Y] & Lancashire wrestling was created in England \\
        P1376 & 216 & [X] is the capital city of [Y] & Ottawa is the capital city of Canada \\
        P1412 & 965 & [X] communicated in [Y] & Lon Chaney communicated in English \\
        \bottomrule
    \end{tabular}
    }
    \caption{All relations we used from ParaRel along with number of items, query template, and an example item for each.}
    \label{tab:pararel-details}
\end{table*}

\begin{figure*}[h!]
    \centering
    \includegraphics[width=\textwidth]{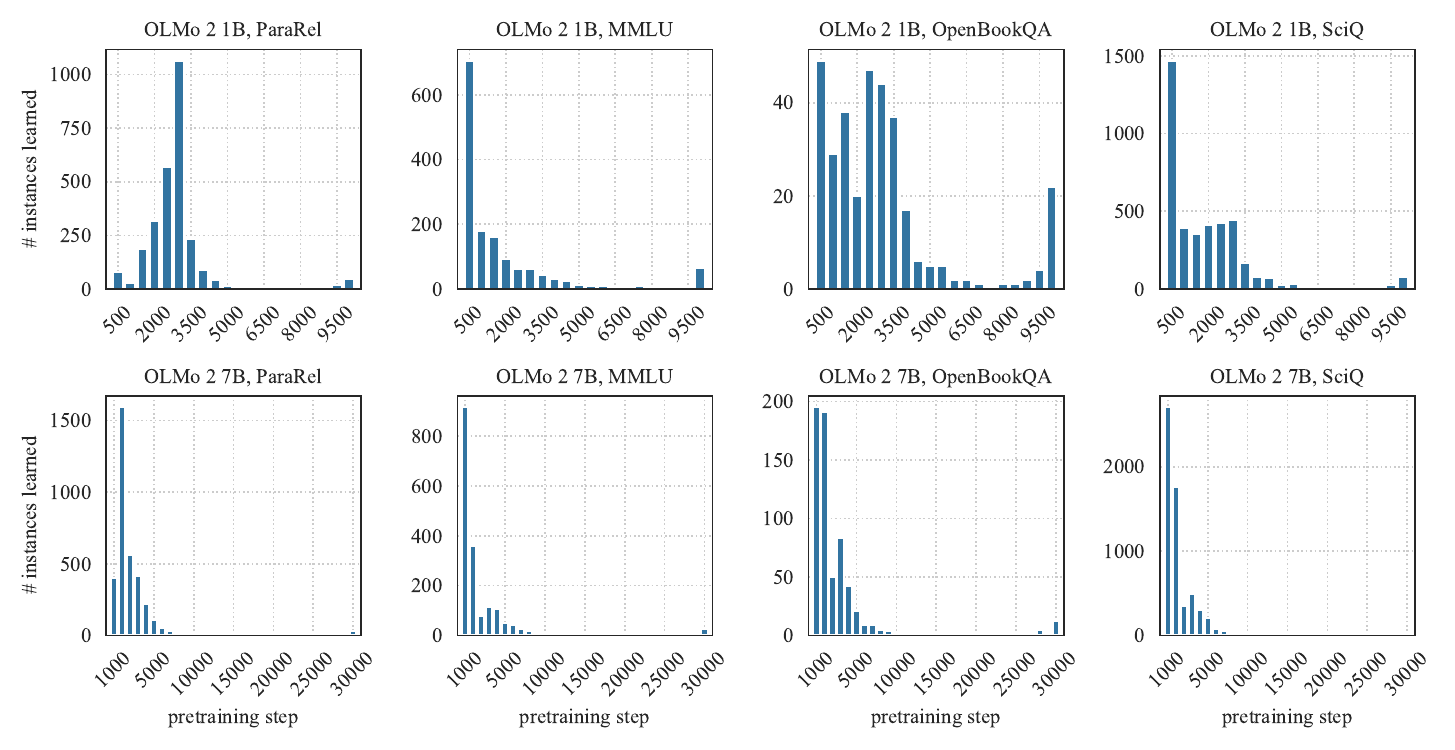}
    \caption{Number of items learned per pretraining step, for each model size and dataset.}
    \label{fig:learned-items}
\end{figure*}

\begin{figure}
    \centering
    \includegraphics[width=0.5\textwidth]{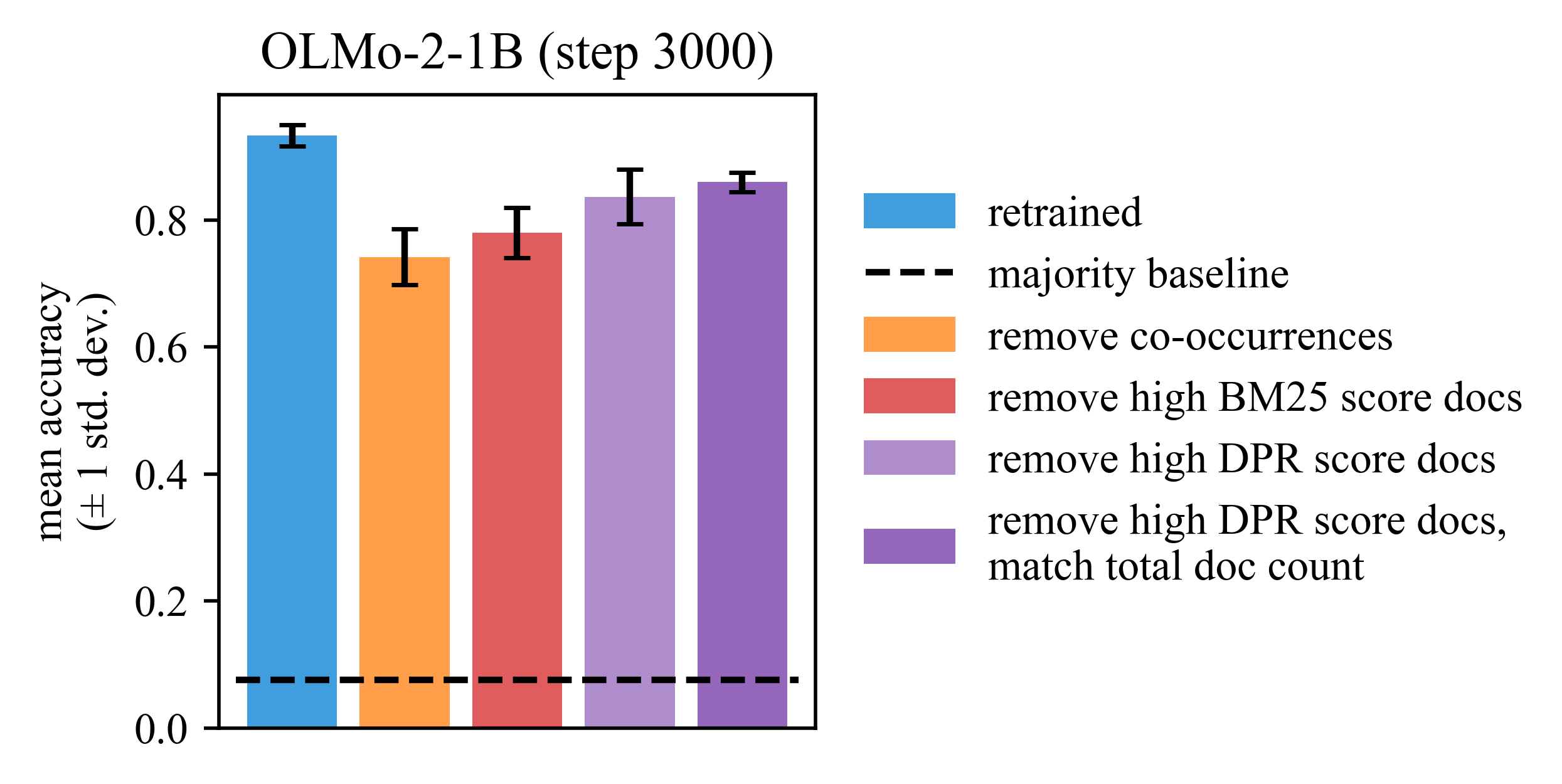}
    \caption{Varying the method of matching documents to items results in different levels of performance reduction on ParaRel items. Selecting relevant documents using BM25 and DPR scores both lead to similar or less reduction in performance compared to selecting based on term cooccurrence alone, when removing the same number of documents per item or in total.}
    \label{fig:vary-matching}
\end{figure}

\begin{figure}
    \centering
    \includegraphics[width=0.5\textwidth]{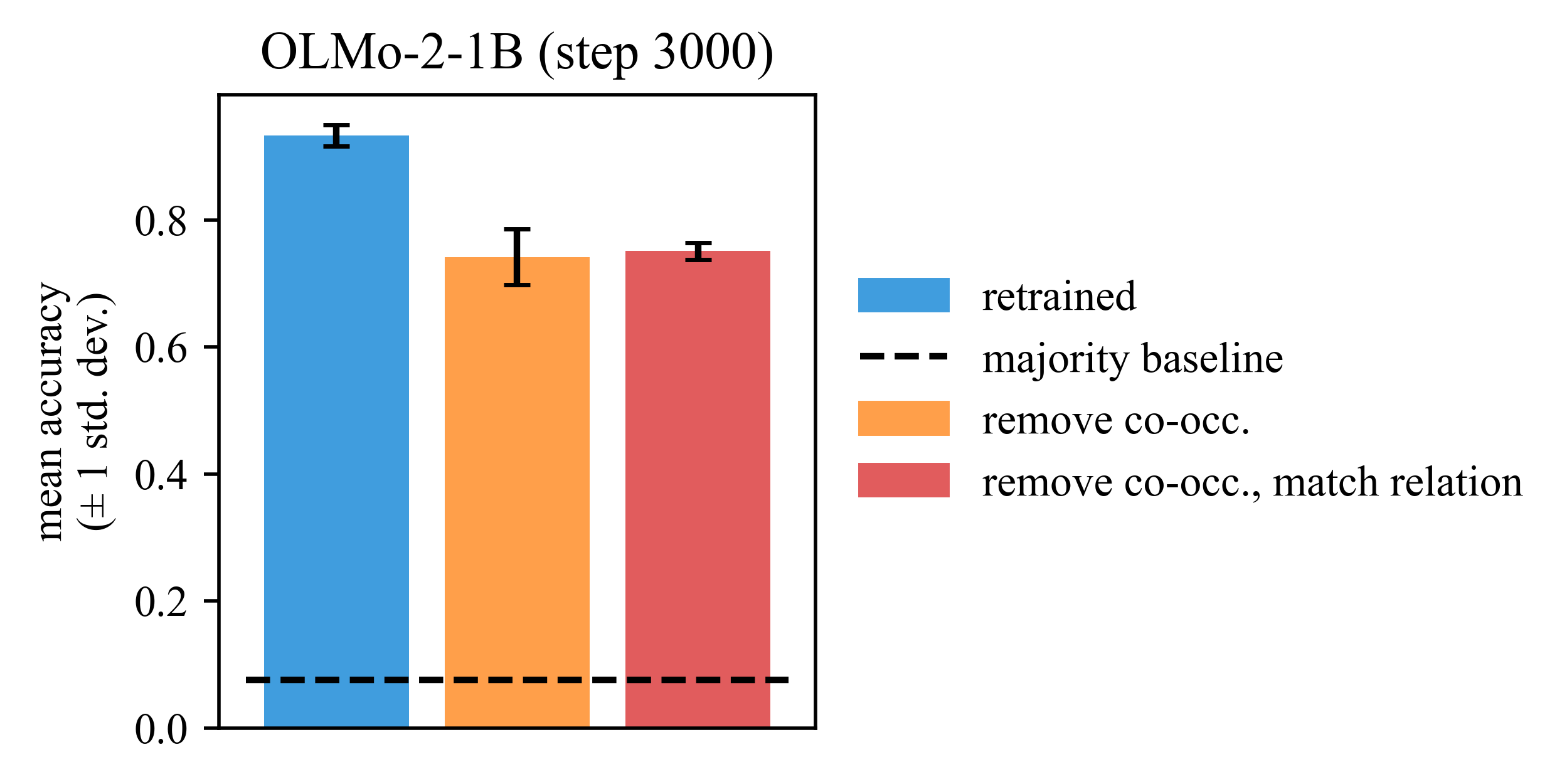}
    \caption{Modifying documents by replacing them with ones expressing the same ParaRel relation type but without having the term cooccurrence has a similar effect to replacing with random unrelated documents, indicating the importance of the specific entity pair being mentioned.}
    \label{fig:vary-modifying}
\end{figure}

\begin{table*}[t!]
    \centering
    \resizebox{\textwidth}{!}{
        \begin{tabular}{l|l|c|l|l}
            \toprule
            Dataset    & Subset(s)               & \# Items & Normalization & Example \\
            \midrule
            \multirow{2}{*}{MMLU}       & \multirow{2}{*}{test}                    & \multirow{2}{*}{14,042} & & The Rhine River is an example of which type of boundary? \\
                                        &                                          &                         & \multirow{-2}{*}{char} & (a) geometric \quad (b) artificial \quad (c) natural \quad (d) relict \\
            \midrule
            \multirow{2}{*}{OpenBookQA} & \multirow{2}{*}{train, validation, test} & \multirow{2}{*}{5,957}  & & What functions without contact between objects? \\
                                        &                                          &                         & \multirow{-2}{*}{PMI} & (a) weight \quad (b) friction \quad (c) opposites attracting \quad (d) pressure \\
            \midrule
            \multirow{2}{*}{SciQ}       & \multirow{2}{*}{train, validation, test} & \multirow{2}{*}{13,679} & & In order to create food, what do photosynthetic protists use? \\
                                        &                                          &                         & \multirow{-2}{*}{char} & (a) decayed matter \quad (b) thermal energy \quad (c) hydrocarbons \quad (d) light energy \\
            \bottomrule
        \end{tabular}
    }
    \caption{Additional details on MMLU, OpenBookQA, and SciQ including subsets used, number of items per dataset, and an example item.}
    \label{tab:mcqa}
\end{table*}

\begin{table*}[t]
    \centering
    \resizebox{\textwidth}{!}{
        \begin{tabular}{l|c|c|c|c|c|c}
        \toprule
        intervention & model size & steps & $\lvert \mathcal{E}^{\text{target}} \rvert$ & \% docs & \% tokens & JSD \\
        \midrule
          & 1B & $1500$--$2000$ & 121 & 2.9\% & 6.1\% & 0.011 \\
          & 1B & $2500$--$3000$ & 367 & 4.7\% & 8.5\% & 0.017 \\
          & 7B & $2000$--$2150$ & 272 & 2.2\% & 4.7\% & 0.010 \\
          \multirow{-4}{*}{suppress learning (coocc.)}
          & 7B & $3000$--$3150$ & 62  & 0.5\% & 1.0\% & 0.003 \\
        \midrule
          & 1B & $1500$--$2000$ & 116 & 24.0\% & 37.3\% & 0.052 \\
          & 1B & $2500$--$3000$ & 362 & 35.0\% & 50.4\% & 0.085 \\
          & 7B & $2000$--$2150$ & 272 & 43.7\% & 53.4\% & 0.096 \\
          \multirow{-4}{*}{suppress learning (coocc.)}
          & 7B & $3000$--$3150$ & 62  & 17.3\% & 28.2\% & 0.047 \\
        \midrule
          & 1B & $1500$--$2000$ & 162 & 3.2\% &  6.3\% & 0.011 \\
          & 1B & $2500$--$3000$ & 88  & 1.3\% &  2.5\% & 0.005 \\
          & 7B & $2000$--$2150$ & 371 & 9.9\% & 18.3\% & 0.031 \\
          \multirow{-4}{*}{suppress learning (coocc.)}
          & 7B & $3000$--$3150$ & 86  & 1.7\% &  3.2\% & 0.008 \\
        \bottomrule
        \end{tabular}
    }
    \caption{Detailed information on the ParaRel interventions for each intervention type (suppressing learning by removing cooccurrences, suppressing learning by removing occurrences, and promoting learning by adding cooccurrences), model size, and set of pretraining steps. Values include the number of target items ($\lvert \mathcal{E}^{\text{target}} \rvert$), percentage of documents changed in the data batch (\% docs), percentage of tokens changed in the data batch (\% tokens), and the Jensen-Shannon divergence between the token distributions of the original and modified data batches (JSD, takes values between 0 and 1).}
    \label{tab:pararel-intervention-stats}
\end{table*}

\begin{table*}[t]
    \centering
    \resizebox{\textwidth}{!}{
        \begin{tabular}{l|c|c|c|c|c|c}
        \toprule
        intervention & model size & steps & $\lvert \mathcal{E}^{\text{target}} \rvert$ & \% docs & \% tokens & JSD \\
        \midrule
          & 1B & $5500$--$6000$ & 82 & 12.1\% & 0.6\% & 0.021 \\
        \multirow{-2}{*}{reduce pro-male bias}
          & 7B & $2000$--$2150$ & 29 &  7.7\% & 0.4\% & 0.014 \\
          & 1B & $5500$--$6000$ & 91 &  7.1\% & 0.7\% & 0.010 \\
        \multirow{-2}{*}{reduce pro-female bias}
          & 7B & $2000$--$2150$ & 18 &  4.5\% & 0.4\% & 0.006 \\
        \bottomrule
        \end{tabular}
    }
    \caption{Detailed information on the WinoBias interventions for each intervention type (reducing pro-male bias, reducing pro-female bias), model size, and set of pretraining steps. Values include the number of target items ($\lvert \mathcal{E}^{\text{target}} \rvert$), percentage of documents changed in the data batch (\% docs), percentage of tokens changed in the data batch (\% tokens), and the Jensen-Shannon divergence between the token distributions of the original and modified data batches (JSD, takes values between 0 and 1).}
    \label{tab:gender-bias-intervention-stats}
\end{table*}

\begin{table*}[t]
    \centering
    \resizebox{\textwidth}{!}{
        \begin{tabular}{l|c|c|c|c|c|c|c}
            \toprule
            dataset & model size & steps & matching method & $\lvert \mathcal{E}^{\text{target}} \rvert$ & \% docs & \% tokens & JSD \\
            \midrule
              &                      &                                  & BM25 &                       & 15.4\% & 18.9\% & 0.038 \\
              &  \multirow{-2}{*}{1B} &  \multirow{-2}{*}{$2000$--$2500$} & DPR  &  \multirow{-2}{*}{424} & 10.5\% & 10.3\% & 0.026 \\
              &                      &                                  & BM25 &                       & 21.3\% & 26.0\% & 0.041 \\
             \multirow{-4}{*}{SciQ}
              &  \multirow{-2}{*}{7B} &  \multirow{-2}{*}{$2000$--$2150$} & DPR  &  \multirow{-2}{*}{353} & 14.4\% & 14.3\% & 0.032 \\
            \midrule            
              &                      &                                  & BM25 &                      & 3.8\%  & 5.1\%  & 0.010 \\
              &  \multirow{-2}{*}{1B} &  \multirow{-2}{*}{$2000$--$2500$} & DPR  &  \multirow{-2}{*}{47} & 3.5\%  & 3.2\%  & 0.008 \\
              &                      &                                  & BM25 &                      & 6.6\%  & 8.4\%  & 0.013 \\
             \multirow{-4}{*}{OpenBookQA}
              &  \multirow{-2}{*}{7B} &  \multirow{-2}{*}{$2000$--$2150$} & DPR  &  \multirow{-2}{*}{50} & 5.6\%  & 5.1\%  & 0.011 \\
            \midrule
              &                      &                                  & BM25 &                       & 10.2\% & 17.9\% & 0.020 \\
              &  \multirow{-2}{*}{1B} &  \multirow{-2}{*}{$1000$--$1500$} & DPR  &  \multirow{-2}{*}{160} & 11.1\% & 10.9\% & 0.014 \\
              &                      &                                  & BM25 &                      & 10.0\% & 16.0\% & 0.020 \\
             \multirow{-4}{*}{MMLU}
              &  \multirow{-2}{*}{7B} &  \multirow{-2}{*}{$2000$--$2150$} & DPR  &  \multirow{-2}{*}{77} & 9.3\%  & 9.7\%  & 0.013 \\
            \bottomrule
        \end{tabular}
    }
    \caption{Detailed information on the MCQA dataset interventions for each dataset, model size, set of pretraining steps, and item-document matching method. Values include the number of target items ($\lvert \mathcal{E}^{\text{target}} \rvert$), percentage of documents changed in the data batch (\% docs), percentage of tokens changed in the data batch (\% tokens), and the Jensen-Shannon divergence between the token distributions of the original and modified data batches (JSD, takes values between 0 and 1).}
    \label{tab:mcqa-intervention-stats}
\end{table*}

\end{document}